# 3D RECONSTRUCTION OF NON-COOPERATIVE RESIDENT SPACE OBJECT USING INSTANT NERF AND D-NERF

Basilio Caruso[*], Trupti Mahendrakar[†], Van Minh Nguyen[‡], Ryan T. White[§], Todd Steffen[**]

The proliferation of non-cooperative resident space objects (RSOs) in orbit has spurred the demand for active space debris removal, on-orbit servicing (OOS), classification, and functionality identification of these RSOs. Recent advances in computer vision have enabled high-definition 3D modeling of objects based on a set of 2D images captured from different viewing angles. This work adapts Instant NeRF and D-NeRF, variations of the neural radiance field (NeRF) algorithm to the problem of mapping RSOs in orbit for the purposes of functionality identification and assisting with OOS. The algorithms are evaluated for 3D reconstruction quality and hardware requirements using datasets of images of a spacecraft mock-up taken under two different lighting and motion conditions at the Orbital Robotic Interaction, On-Orbit Servicing and Navigation (ORION) Laboratory at Florida Institute of Technology. Instant NeRF is shown to learn high-fidelity 3D models with a computational cost that could feasibly be trained on on-board computers.

## INTRODUCTION

The demand for space domain awareness (SDA) has been on the rise with the increasing amount of spacecraft and debris in orbit. With this rapid proliferation, identification of the resident space objects (RSO) is a crucial task for determining the type of the RSO for On-Orbit Servicing (OOS), contingency, and space debris removal applications. Various methods have been proposed ranging from ground observation [1, 2] to in-space on-orbit observations [3, 4].

Our work at the ORION Facility [5] at Florida Institute of Technology focuses on autonomously navigating around RSOs with Distributed Satellite Systems (DSS) while in-orbit [6]. Many RSOs in orbit are large, non-cooperative, lack navigational aids and capture interfaces, and could be structurally damaged and/or tumbling. A highly scalable solution for performing Active Debris Removal (ADR) and OOS of such large spacecraft would be to use DSS, which would significantly reduce the forces and moments acting on the capture interfaces.

OOS and ADR operations have been demonstrated many times with the space shuttle operations. Some missions include Palapa-B2, Westar VI, Hubble Space Telescope, and the ISS [7].

---

[*] Master's Student, Flight Test Engineering, Florida Institute of Technology, 150 W University Blvd, Melbourne, FL.
[†] PhD Candidate, Aerospace Engineering, Florida Institute of Technology, 150 W University Blvd, Melbourne, FL.
[‡] PhD Candidate, Mathematical Sciences, Florida Institute of Technology, 150 W University Blvd, Melbourne, FL.
[§§] Assistant Professor, Mathematical Sciences, Florida Institute of Technology, 150 W University Blvd, Melbourne, FL.
[**] Engineer, Technetium Engineering LLC, 7901 4th St N, STE 300, St. Petersburg Fl 33702.

However, with the increasing costs and risks associated with crewed missions, the focus of OOS shifted to robotic on-orbit servicing. Some missions that have demonstrated robotic OOS are ETS-VII [8] in 1997, DART [9], XSS-10/-11 [10, 11], MiTEX [12], ANGELS [13], MEV-1/2 [14, 15] and ELSA-d [16].

The above missions show the commitment of the space industry to find solutions enabling ADR and OOS. All these missions used a custom robotic servicer built for cooperative RSO servicing operations. Close proximity operations with non-cooperative RSOs are much more difficult due to greater uncertainties in pose, rotation rates, and structural conditions of the RSO. Compounding the complexity of such missions, human-in-the-loop operations are infeasible due to time delay in on-board telemetry. These challenges demand fully autonomous missions, but fully autonomous operations capable of generating safe approach and capture trajectories to a non-cooperative RSO have not been successfully executed yet.

Robust inspection capabilities to infer the geometry and features of the RSO are prerequisites for reliable autonomous OOS operations with non-cooperative RSOs. In particular, in-space inspection can enable the development of intelligent path planning algorithms. Deep learning-based computer vision algorithms (perhaps enhanced with sensor fusion and swarms) show the greatest potential to solve these problems. While the computational cost for deep learning methods are relatively high, recent innovations have been reducing computational costs while on-board spacecraft hardware computational capabilities continue to improve with faster, more energy efficient computers.

Previous research conducted in [3, 6, 17] demonstrated successful classification and localization of different potential capture points and fragile components of the RSO, including thrusters, antennas, solar panels and satellite bodies through state-of-the-art computer vision and sensor fusion techniques. However, these works do not map the 3D structure of the RSO with components other than the ones mentioned above. This means the chasers are still prone to the risk of collision near an unfamiliar RSO.

The present work focuses on the problem of creating a high-fidelity 3D model of the target RSO based on a set of static images of the RSO that may be captured from an inspection orbit. Work by the European Space Agency (ESA) [18] demonstrated deep learning-based view synthesis algorithms can be used to infer the 3D geometry of an RSO prior to executing a capture trajectory. The ESA paper compares the quality of 3D models learned from still images of CAD models using the neural radiance field (NeRF) [19] and generative radiance field (GRAF) [20] algorithms trained on 2D scenes of fixed CAD models captured from many angles.

This paper goes a few steps further to answer three key research questions:

1. Can NeRF-related algorithms learn high-fidelity 3D models based on 2D scenes of a real-life satellite mock-up under space-like conditions?

2. Can an accelerated version of NeRF produce high-fidelity 3D models at a far lower computational cost and how does it relate to on-board computing capabilities?

3. Is an algorithm designed for dynamic rather than static 3D objects required to produce high-fidelity 3D models of satellites?

To address these questions, NeRF-related algorithms are used to build 3D models. The training data (2D scenes as RGB images) are captured through hardware-in-the-loop experiments, where images of a real-life satellite mock-up are captured using a kinematics simulator in the ORION Facility at Florida Tech [5]. In addition, our target RSO will be in motion and the algorithms will be tested under different lighting conditions.

Initial experiments confirmed both NeRF and instant neural graphics primitive (NGP)-accelerated NeRF (referred to as Instant NeRF) [21] produced 3D models that generate high-quality novel 2D scenes. There are some artifacts in the computer-generated scenes, but this is mostly limited to minor noise or blurry reconstructions of highly reflective surfaces. Neither of these are big problems for inferring 3D geometries of target RSOs. Instant NeRF trains more quickly than NeRF by a factor of 300-400 with little to no difference in visual quality. Instant NeRF could be trained on small, on-board computers in 6-7 minutes. (These figures are based on NVIDIA Jetson TX2, which was used on a satellite launched in 2022 [22].) Dynamic NeRF (D-NeRF) [23] is a variation of the NeRF algorithm designed to operate on dynamic 3D objects, but it is shown in the experiments below to be unnecessary, and seemingly less suited to the dynamic objects captured in our experiments for this application.

The next section is a discussion of NeRF and related algorithms in more depth. The following section contains specifications of the experimental, the approach and challenges of data collection performed in the study, and details the metrics used to measure the quality of 3D models. The section after presents the results of the algorithms – Instant NeRF, D-NeRF and NeRF, lighting, and motion comparisons, including novel views of the satellite generated by the 3D models. The last section provides conclusions and future work.

**NERF AND RELATED ALGORITHMS**

NeRF [19] is based on the geometrical optics approximation, which makes a simplifying assumption that light travels along simple straight-line paths (rays) rather than via wave propagation. This assumption is widely used in computer graphics because it allows computers to store 3D models based on a representation of how light emanates and is absorbed by points in space–these structures are called radiance fields. A high-quality radiance field can be used to render a scene of the 3D object from any chosen viewing angle on demand using ray tracing to trace a dense set of rays of light emanating from the object to the viewer. This provides a scene of the model from the given viewing angle.

While NeRFs can be trained in a few hours on a single modern GPU on ground, they are not suitable for training with on-board spacecraft hardware due to the computational load because these representations are large, fully connected neural networks (8 layers with 256 neurons each followed 281-neuron and 128-neuron layers). Instant neural graphics primitives [21] provide a learnable approach to encode the input data such that a much smaller neural architecture (3 layers with 64, 80, and 64 neurons, respectively) can be jointly trained without substantial loss of model fidelity. This acceleration reduces the computational cost by several orders of magnitude, nearly enabling on-board training with current NVIDIA Jetson computers that have been used in space in recent years [22].

NeRF and Instant NeRF showed good potential in initial experiments, but these algorithms were designed for 3D modeling of static objects. It was not clear how well they would work under the changing attitudes and lighting conditions an orbiting RSO may experience while being observed. The recently introduced Dynamic NeRF (D-NeRF) [23] explicitly considers rigid and non-rigid motion of the target in constructing a 3D model based on a camera moving around the object. It does this by training two deep neural networks. One learns to encode 2D scenes into a canonical representation. The second network learns to map these canonical representations to deformed scenes based on a time input. The two networks can learn high-fidelity 3D models with a 2.5 times higher compute cost than NeRF but with an extra input of time to deal with the motion. While there is no equivalent of Instant NGP-acceleration of D-NeRF, good performance by D-NeRF in these experiments would motivate developing such an algorithm.

**EXPERIMENTAL SETUP**

Experiments were performed at the ORION facility at Florida Tech [5] to capture images of a mock-up of a satellite to represent an RSO. The ORION facility is equipped with a planar Maneuver Kinematics 2DOF Motion table as shown in the figure below. The simulator can translate and rotate a payload at a maximum speed of 0.25 m/s at a maximum acceleration of 1 m/s$^2$ along both linear axes. The pan-title mechanisms are specifically designed to hold a test article that weights up to 20 kgs and has dimensions of 0.5 m × 0.5 m × 0.5 m. The mechanisms have a range of motion in elevation of ±90° and can rotate infinitely in the azimuth direction, with a maximum rotation rate of 60°/s and a maximum acceleration of 60°/s$^2$ about each axis.

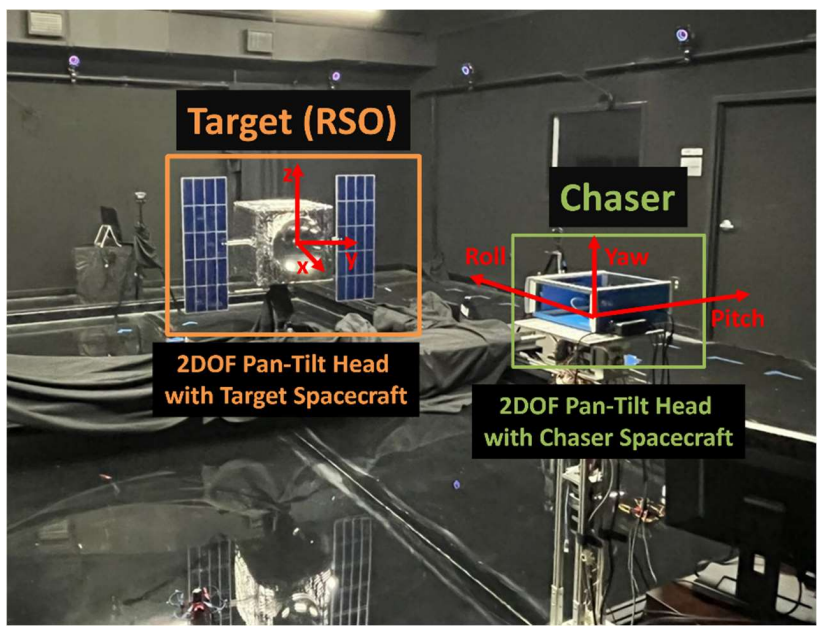

**Figure 1: ORION Simulator**

The lighting conditions in the facility were simulated by a Hilio Light Panel with a color temperature of 5600 K rated for 350 W of power with adjustable intensity. When powered, the maximum intensity of the panel was equivalent to a 2000 W incandescent lamp. For the experiments performed in this paper, the intensity of the lamp was set to 10% and 100% in different experiments to simulate extreme case scenarios. Lighting under 10% yielded intermittent to no visual of rotating spacecraft. Hence, a minimum threshold of 10% was chosen.

In each experiment below, images of the target RSO were captured by a single monocular camera (iPhone 13) mounted to a chaser were used as inputs to the Instant NeRF and D-NeRF algorithms.

**Data Collection**

A total of four cases were evaluated. The goal is to study the feasibility of Instant NeRF and D-NeRF in comparison to NeRF to answer the three key research questions above.

**Case 1.** Images of the target RSO are taken at 10° increments around a circle of diameter of 5 meters (simulating an R-bar maneuver around a stationary RSO) with 10% lighting intensity. (See Figure 2.)

**Case 2.** Images of the target RSO are taken at 10° increments around a circle of diameter of 5 meters (simulating an R-bar maneuver around a stationary RSO) with 100% lighting intensity. (See Figure 3.)

**Case 3.** Videos of the RSO are captured as it yaws at 10/s with the chaser positioned at 5 feet (simulating V-bar station keeping around a spinning RSO) with 10% lighting intensity.

**Case 4.** Videos of the RSO are captured as it yaws at 10°/s with the chaser positioned at 5 feet (simulating V-bar station keeping around a spinning RSO) with 100% lighting intensity.

For all four cases, the lamp was pointing directly at the mock-up. The camera aperture was positioned 5 ft from the target spacecraft at a height of 3 feet 8 inches from the ground. Throughout the duration of test data accumulation, the lighting conditions were fixed. Each image and video is scaled to resolution 591-by-443 and the videos use a 0.5 Hz framerate. The framerate is set intentionally low to ensure the 3D model construction is suitably difficult for realistic inspection scenarios, where suitable 2D image captures may be sparse.

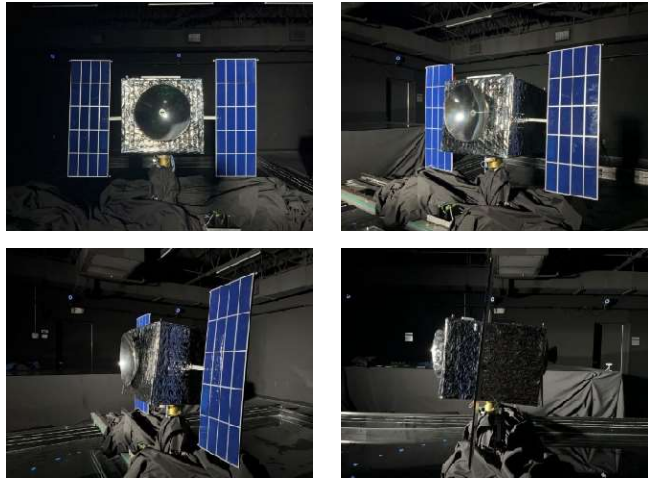

**Figure 2: Case 1 - 10% Brightness**

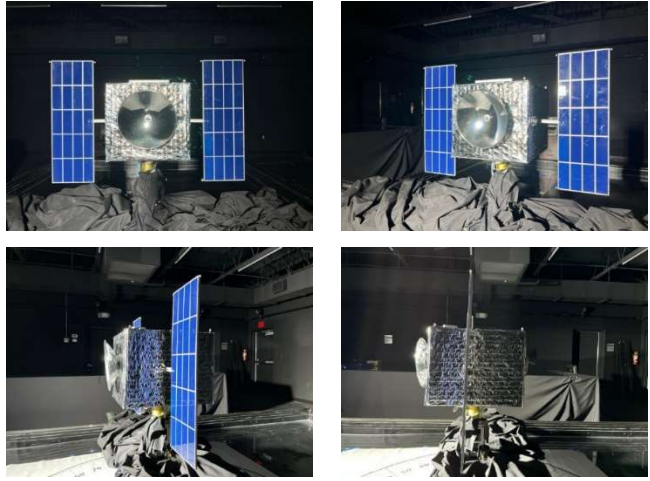

**Figure 3: Case 2 - 100% Brightness**

The 3D reconstruction pipeline requires camera pose of each images used. To extract camera pose for our implementation, we used a popular toolkit known as COLMAP [24]. COLMAP camera pose estimation consists of 3 steps: feature matching, geometric validation, and sparse reconstruction. As the dataset used for this research is a collection of images / image frames extracted from a video, sequential matching was used for feature matching and geometric validation. With the exported scene graph from feature matching, we used a simple radial camera model (assuming all images have different camera calibration) to reconstruct the first pass of 3D model with the camera pose. However, we could have chosen exhaustive matching but it is computationally expensive. **On the other hand, we did not investigate pinhole camera model which could improve the camera pose result as we assume a consistent camera priori.**

In initial experiments, we observed the COLMAP script was not able to map the camera locations because the script anchored the images to the stationary background in the lab (e.g., walls, floor, corners) when the camera was stationary and the satellite was spinning, the COLMAP. To work around this issue with the COLMAP, we tried three approaches. First, we modified the COLMAP transform file template to generate camera coordinate anchor points. Since COLMAP reconstructs the 3D object with the assumption that the object was stationary while camera was moving (Structure-from-Motion), inferences from the sparse COLMAP models did not support the hypothesis of Dynamic NeRF, where both the camera and object were moving. Instead of using COLMAP to infer the camera pose, it was assumed the current camera pose to be at canonical form (identity transformation, a 1x2 block matrix with first block being 3x3 identity matrix, second block was 3x1 zero vector). As shown Figure 4, this method did not yield any distinguishable 3D render of the target. Further attempt was by removing the background with U-Net, in the hope the model did not get confused by background texture, but the result did not improve.

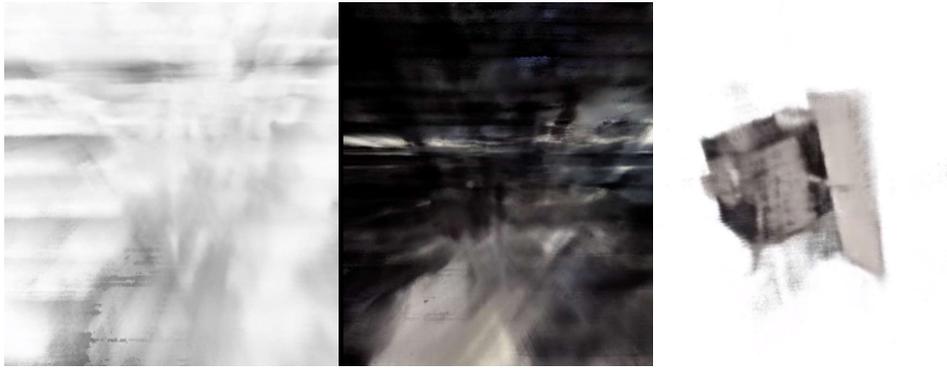

**Figure 4. D-NeRF Attempts to render the Spinning Satellite Target**

In tandem with the first approach, we also tried using LIDAR sensor from iPhone 13 and ARKit to generate a proper camera pose. However, as the camera was stationary, there was no ray/pose change, resulting in a non-invertible ray change matrix and it was not possible to generate transformation matrix from the data. It was concluded that using D-NeRF does not automatically resolve the target relative rotation issue, and it still necessitates the use of mapping software and an estimated camera pose/transformation file. With the information learned from the first approach, we needed to rethink our approach to data collection methodology. To trick COLMAP into thinking the camera is rotating around the object, the next approaches were explored.

Second, we eliminated the target satellite's background using the Intel Realsense D435i stereoscopic depth camera. The idea was to record the motion of the satellite while ignoring points further away in the background. Figure 5 shows the results from post processing the footage from the depth camera. But the camera fails to generate a mask with fine details. The borders are not crisp and the whole object is distorted. Additionally, it was also noticed that under lower lighting conditions and larger distance from the object, the camera tends to lose pixel information in the frame causing it to output black boxes in certain regions. This method was deemed not effective for the purpose of generating accurate preprocessed input images for Instant NeRF or D-NeRF.

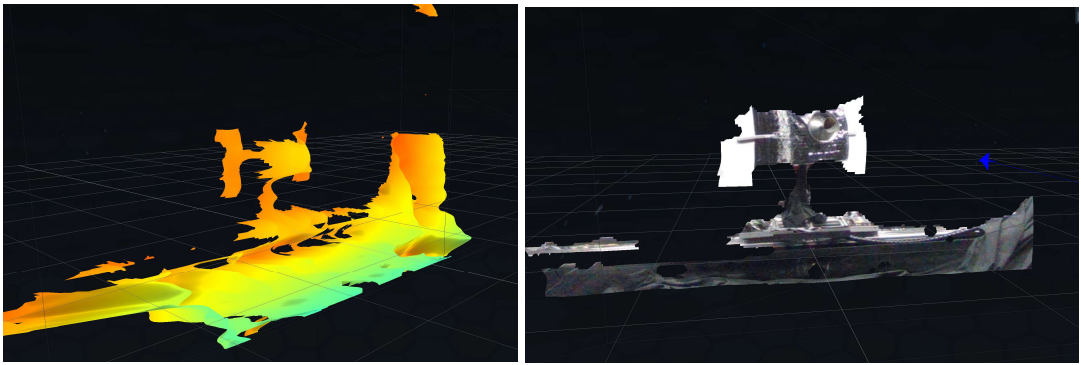

**Figure 5. Depth Camera Captures from RealSense Viewer**

Third, we used a green screen behind the spacecraft to eliminate the background. Green screens are commonly used in post-production to remove background and add special effects in a technique called chroma key compositing. It is a technique for layering (or compositing) images or video streams in regions with specified color ranges (e.g. the green screen). We used this technique to remove the background of the target as it yawing at 10°/s. Figure 6 shows the results from this third approach. As seen below, chroma keying successfully eliminated the background of the input image while still preserving the edges, texture, and shape features of the target satellite.

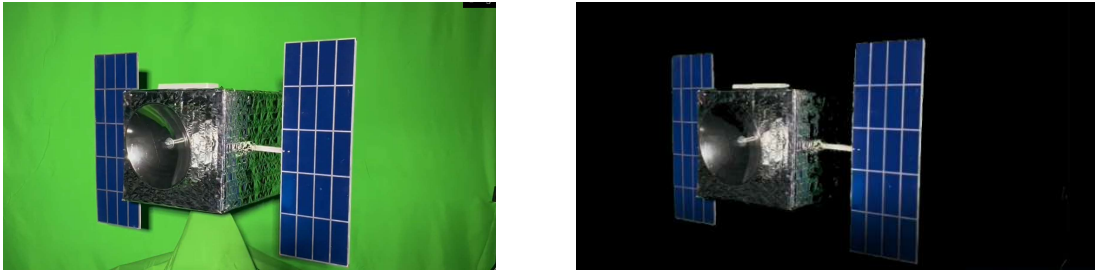

**Figure 6. Chroma Key Compositing**

Once the target object was isolated from the background in the videos using chroma keying, images from the videos at a rate of 2 frames per second were extracted. A total of 80 input images were generated from each video. These images were then input to COLMAP. COLMAP was able to map and anchor all 80 camera views, shown in Figure 8 to generate a transform.json file with a 10% lighting condition setting (Case 3).

An attempt to generate dataset at 100% lighting condition (Case 4) was made, but due to the difference in tonality of the prominent shadow of the body over the green screen, instant NeRF did not train accurate renders. The problem was not further investigated since the purpose of the green screen experiment was to validate mapping of the input pictures under the relative motion of the target satellite with respect to the chaser satellite.

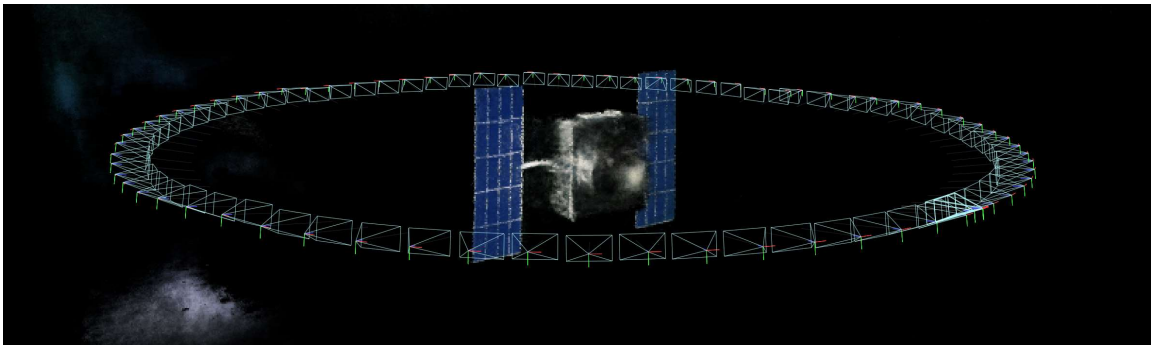

**Figure 7. Camera Visualization**

### 3D Modeling

3D models were constructed under each set of lighting conditions with the Instant NeRF, D-NeRF, and NeRF algorithms.

### Model Quality Metrics

Peak to Signal Noise Ratio (PSNR), Structural Similarity Index (SSIM) and Learned Perceptual Image Patch Similarity (LPIPS) were used to evaluate the performance of Instant NeRF, D-NeRF and NeRF algorithms. The PSNR is a measure of performance of the rendered image at a pixel level. Higher the PSNR value, better is the performance. The SSIM is a measure of the difference between the original picture and the reconstructed image that accounts for changes such as luminance or contrast [25]. Higher the SSIM value, better is the performance. LPIPS is a measure of image similarity between the image patches the algorithm generates and the real image of the object [26]. Lower LPIPS value indicates better performance of the algorithm.

## RESULTS

The results from the Instant NeRF, D-NeRF in comparison with NeRF are tabulated below.

### Table 1: Results

| Scene | Algorithm | PSNR | SSIM | LPIPS |
|---|---|---|---|---|
| Case 1<br>*10% lighting<br>moving camera* | NeRF | 25.866 | 0.722 | 0.463 |
| | Instant NeRF | 21.966 | 0.712 | **0.238** |
| | D-NeRF | **26.111** | **0.732** | 0.448 |
| Case 2<br>*100% lighting<br>moving camera* | NeRF | **24.528** | 0.743 | 0.429 |
| | Instant NeRF | 24.133 | **0.841** | **0.212** |
| | D-NeRF | 8.442 | 0.023 | 0.684 |
| Case 3<br>*10% lighting<br>stationary camera<br>green screen* | NeRF | **26.768** | 0.893 | 0.140 |
| | Instant NeRF | 23.874 | **0.925** | **0.081** |
| | D-NeRF | 14.875 | 0.795 | 0.194 |

For the Case 1 experiment, D-NeRF outperformed both Instant NeRF and NeRF in PSNR and SSIM. Case 2 with 100% lighting, D-NeRF could not generate proper reconstruction of the satellite model. This could be one of the downsides of the temporal network presented by D-NeRF and needs further investigation. Additionally, D-NeRF failed to perform well with in Case 3.

Overall, for all 3 cases, both NeRF and Instant NeRF performed well and were able to reconstruct 3D model of the satellite. While Instant NeRF fell behind in standard image reconstruction metrics (PSNR, SSIM), it excels in "perceived" human judgment (LPIPS score), widely considered a superior metric in the computer vision community. As such, the result generated (as shown in images below) looks more realistic. Also, Instant NeRF has a significant advantage in computational cost as shown in the runtime table below. Both NeRF and D-NeRF required a higher-powered GPU and longer runtime unlike the more efficient Instant NeRF. Each method was trained until it converged (50000 epochs for NeRF and D-NeRF and 35000 epochs for Instant NeRF).

### Table 2: Runtime Metrics

| Algorithm | Device | Average Runtime |
|---|---|---|
| NeRF | RTX 3080 Ti | 70m |

| Instant NeRF (NGP) | RTX 2060 | 15m |
|---|---|---|
| D-NeRF | RTX 3080 Ti | 67m |

3D render results for Cases 1, 2 and 3 from Instant NeRF are shown below. Note the center image is a ground-truth image while the nine surrounding images in each case are novel views that are *not* in the training dataset. Rather, the algorithms learn to generate them based on the radiance field they learn.

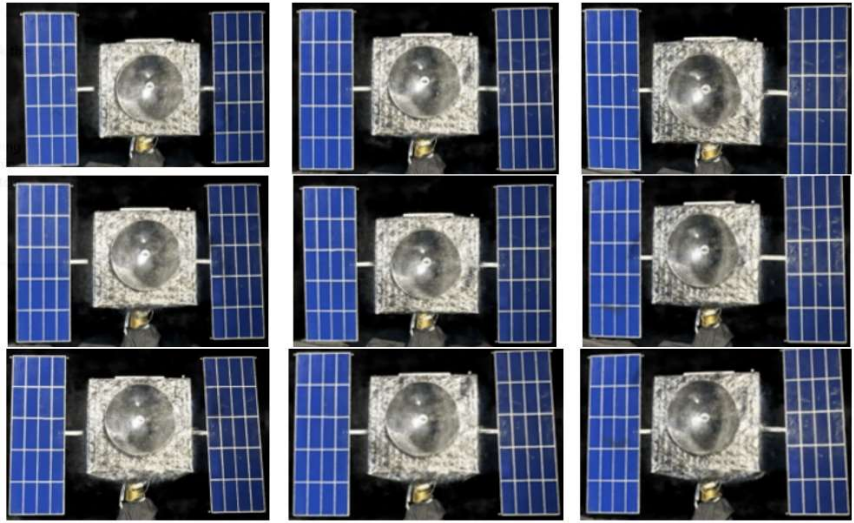

**Figure 8: Case 1: 3D Model Render under 10% Lighting**

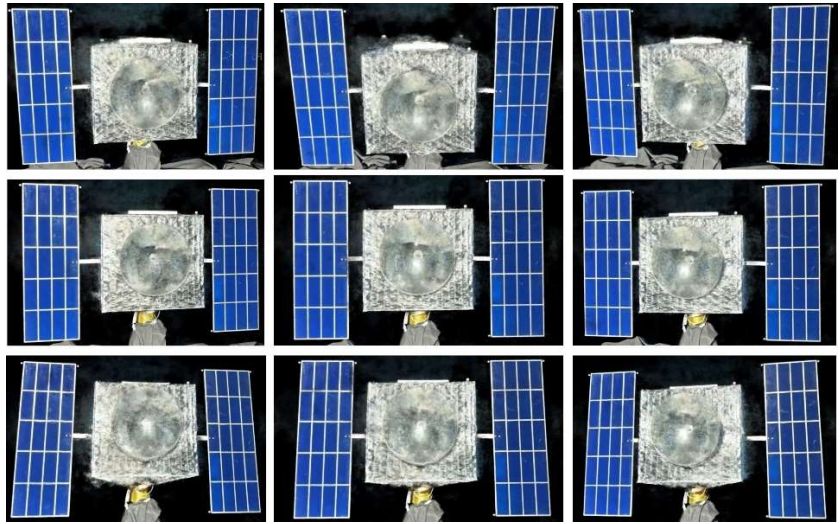

**Figure 9: Case 2: 3D Model Render Under 100% Lighting**

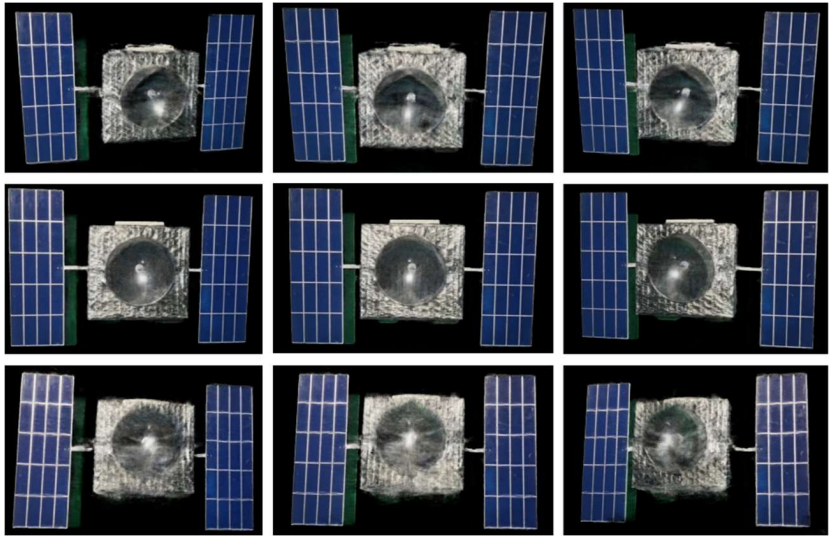

**Figure 10: Case 3: 3D Model Render with the Green Screen Under 10% Lighting**

## FUTURE EXPERIMENTS AND CONCLUSION

The lower computational requirements, much faster runtime along with relatively lower LPIPS value of Instant NeRF makes it a more favorable algorithm compared to NeRF and D-NeRF. Additionally, demonstrating hardware-in-the-loop experiments to evaluate the algorithms proved that though there are drawbacks with each algorithm 3D reconstruction of a spacecraft under realistic lighting conditions with all three algorithms are possible.

Successful rendering of Case 3 also proves that the algorithms are not only capable of producing 3D models of a static object but also when the object is spinning as long as we are able to precisely remove the background. This implies a chaser satellite along the V-bar of a spinning target satellite is still able to produce a 3D render of the target.

For Cases 1 and 2, we assumed a simple radial camera model: i.e., with the chaser revolving around the target at a radial distance of 5m. This type of model assumes that the camera calibration remains unchanged, but this assumption would only work if the light source and the target satellite are relatively stationary. This assumption constrains the approach for our application, but implementing a pinhole camera model will resolve this issue as we know that the real camera in that model is stationary.

Future work will evaluate the Instant NeRF algorithm with different inspection orbits such as R-bar, V-bar, corkscrew, and teardrop on synthetic and hardware-in-the-loop images of a tumbling RSO under varying lighting conditions. The algorithm will also be evaluated on NVIDIA Jetson hardware and will be added to the artificial potential field guidance algorithm developed in [6, 27] to enable fully autonomous on-orbit servicing experiments AI-based feature recognition [3, 6, 17]. All this research together will form the foundation to enable autonomous swarm satellite operations using artificial potential field guidance algorithms around a non-cooperative RSO.